\documentclass{article}
\usepackage{arxiv}
\usepackage[utf8]{inputenc} 
\usepackage[T1]{fontenc}    
\usepackage{hyperref}       
\usepackage{url}            
\usepackage{booktabs}       
\usepackage{amsfonts}       
\usepackage{nicefrac}       
\usepackage{microtype}      
\usepackage{graphicx}
\usepackage{doi}

\usepackage{amsmath}
\usepackage{subcaption}
\usepackage{multicol}
\usepackage{multirow}
\usepackage[table,xcdraw]{xcolor}
\usepackage{threeparttable}
\usepackage{ulem}
\usepackage{verbatim} 

\begin{document}
\title{Hilbert curves for efficient exploratory landscape analysis neighbourhood sampling}
%
%

\author{ Johannes J. Pienaar\\
	Department of Computer Science\\
	University of Pretoria \\
        Pretoria, South Africa \\
	\texttt{jpienaar85@gmail.com} \\
	\And
	\href{https://orcid.org/0000-0003-3546-1467}{\includegraphics[scale=0.06]{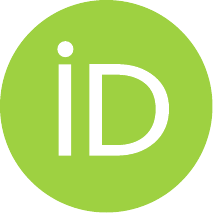}\hspace{1mm}Anna S. Bosman} \\
	Department of Computer Science\\
	University of Pretoria \\
        Pretoria, South Africa \\
	\texttt{anna.bosman@up.ac.za} \\
 \And
	\href{https://orcid.org/0000-0002-6070-2632}{\includegraphics[scale=0.06]{orcid.pdf}\hspace{1mm}Katherine M. Malan} \\
	Department of Decision Sciences\\
	University of South Africa \\
        Pretoria, South Africa \\
	\texttt{malankm@unisa.ac.za} \\
}
\maketitle              
\begin{abstract}
Landscape analysis aims to characterise optimisation problems based on their objective (or fitness) function landscape properties. 
The problem search space is typically sampled, and various landscape features are estimated based on the samples.
One particularly salient set of features is \textit{information content}, which requires the samples to be sequences of neighbouring solutions, such that the local relationships between consecutive sample points are preserved. Generating such spatially correlated samples that also provide good search space coverage is challenging. It is therefore common to first obtain an unordered sample with good search space coverage, and then apply an ordering algorithm such as the nearest neighbour to minimise the distance between consecutive points in the sample. However, the nearest neighbour algorithm becomes computationally prohibitive in higher dimensions, thus there is a need for more efficient alternatives.
In this study, Hilbert space-filling curves are proposed as a method to efficiently obtain high-quality ordered samples. Hilbert curves are a special case of fractal curves, and guarantee uniform coverage of a bounded search space while providing a spatially correlated sample. We study the effectiveness of Hilbert curves as samplers, and discover that they are capable of extracting salient features at a fraction of the computational cost compared to Latin hypercube sampling with post-factum ordering. Further, we investigate the use of Hilbert curves as an ordering strategy, and find that they order the sample significantly faster than the nearest neighbour ordering, without sacrificing the saliency of the extracted features.

\end{abstract}
\section{Introduction}
Search landscape analysis has established itself as a useful approach for understanding complex optimisation problems and analysing evolutionary algorithm behaviour~\cite{Malan2021}. Many landscape analysis techniques have been developed over the years, with the most widely used techniques including fitness distance correlation~\cite{Jones1995}, local optima networks~\cite{OCHO2008}, and exploratory landscape analysis (ELA)~\cite{Mersmann2011}. When landscape analysis produces numeric outputs, the resulting feature vectors can be used as abstract representations of problem instances, where instances with similar feature vectors are assumed to fall into similar problem classes. If these feature vectors effectively capture the important characteristics of problems, they can be used as the feature component for automated algorithm design (AAD) -- specifically, automated algorithm configuration and selection. A number of recent studies have achieved different aspects of AAD using landscape analysis in specific contexts~\cite{Beham2018,Kostovska2023,KUK_2019,Liefooghe2023,MALA2018,SALL2020}. 

Landscape analysis approaches differ in terms of what they measure or predict (e.g. ruggedness, modality, presence of funnels, and so on), and also on what they produce (e.g. numerical results or a visualisation of a phenomenon)~\cite{Malan2013}. They can also be distinguished based on the scale of the analysis~\cite{Pitzer2012} -- a global approach attempts to characterise the features of the search space as a whole, while a local approach will consider the features of the landscape in the neighbourhood of solutions. Fitness distance correlation~\cite{Jones1995} and local optima networks~\cite{OCHO2008} are both examples of global approaches, whereas the average length of an adaptive walk~\cite{Verel2013} is an example of a local landscape feature. 

Many landscape analysis techniques that measure local features are based on samples that are {\it spatially correlated}, i.e. sequences of neighbouring solutions, as opposed to a sample of independent solutions from the whole search space.  Techniques that require such sequences of sampled solutions include correlation length for measuring ruggedness~\cite{Weinberger90}, entropic profiles of ruggedness and smoothness with respect to neutrality \cite{Vassilev2000,Vassilev2003} -- adapted as a single measure of ruggedness for continuous spaces~\cite{Malan2009}, approximations of gradient~\cite{Malan2013a}, information content features~\cite{Munoz2015_ELA}, and measures of neutrality~\cite{Aardt2017}. 

In the context of numerical optimisation, a wide range of landscape analysis approaches have been implemented in the R package \texttt{flacco}~\cite{Kerschke2019}, which has also been re-implemented in Python (\texttt{pflacco}\footnote{\url{https://pypi.org/project/pflacco/}}). Over 300 landscape metrics are included in \texttt{flacco}, organised into 17 sets, which include the original ELA metrics~\cite{Mersmann2011} covering six of the feature sets, and the set of information content features~\cite{Munoz2015_ELA} consisting of five metrics. In a study of the subset of the ``cheap" feature sets in \texttt{flacco}, Renau et al.~\cite{Renau2019} found that the two most salient (in terms of distinguishing between problems) and robust feature metrics were from the information content and ELA meta-model feature sets. Unlike the other feature sets in \texttt{flacco}, the information content metrics require spatially correlated samples.
 The saliency of the information content feature metrics and the additional requirement of neighbourhood ordering is the motivation for this study.  

To date, the two main strategies proposed for generating spatially correlated samples in continuous search spaces are random walks~\cite{Malan2014} and post-factum ordering of global samples~\cite{Munoz2015_ELA}. Desirable properties of a sampling strategy for estimating local features are that the solutions provide good coverage of the search space~\cite{Lang2020a}, successive points are positioned close to each other (compared to other solutions in the sample) to capture landscape changes in the neighbourhood, and that the process has low computational expense. The most commonly used approaches to neighbourhood sampling for numerical optimisation are uniform random sampling or Latin hypercube sampling (LHS)~\cite{Renau2020}, followed by either random or nearest-neighbour ordering. These are the approaches implemented in \texttt{flacco} and \texttt{pflacco}.

In this paper, we investigate the use of Hilbert curves~\cite{Hilbert1935} as a new sampling strategy. Hilbert curves are fractal space-filling curves with two desirable properties: (1) they guarantee uniform search space coverage, and (2) neighbouring points on the curve are located close to one another. We show that Hilbert curves are comparable to ordered LHS in terms of the saliency of the extracted features, but are significantly cheaper to compute. Additionally, we show that a Hilbert curve can be used to spatially order an LHS sample, resulting in significant computational gains compared to the commonly used nearest neighbour ordering method.   



\section{Hilbert curves}\label{sec:space-filling}
A space-filling curve is a surjective continuous function from the unit interval $[0,1]$ to a unit hypercube $[0,1]^d$. Surjectivity implies that every point in the hypercube maps to at least one point in the interval, and continuity ensures that no areas in $[0,1]^d$ are missed~\cite{Falconer2004}. Space-filling curves are a special case of fractal curves, and are guaranteed to fill a continuous space in the limit. 

Our interest in space-filling curves derives from two useful properties that they offer, namely (1) uniform coverage of a bounded $d$-dimensional space, and (2) the ability to provide a unique mapping between points in the $d$-dimensional space and points on the  1-dimensional curve~\cite{sagan2012space}. Specifically, this study considers the Hilbert space-filling curve, first proposed by D. Hilbert in 1891~\cite{Hilbert1935}. The Hilbert curve is defined recursively, and can be constructed through a limit process of iteration. Each successive iteration creates an approximation of the true Hilbert curve that passes through more points in the $d$-dimensional unit hypercube. For practical purposes, the number of iterations is chosen to be finite, and is further referred to as the \textit{order} of the Hilbert curve. 

\begin{figure}[!tbp]
    \centering
    \begin{subfigure}{0.32\textwidth}
        \includegraphics[width=\linewidth]{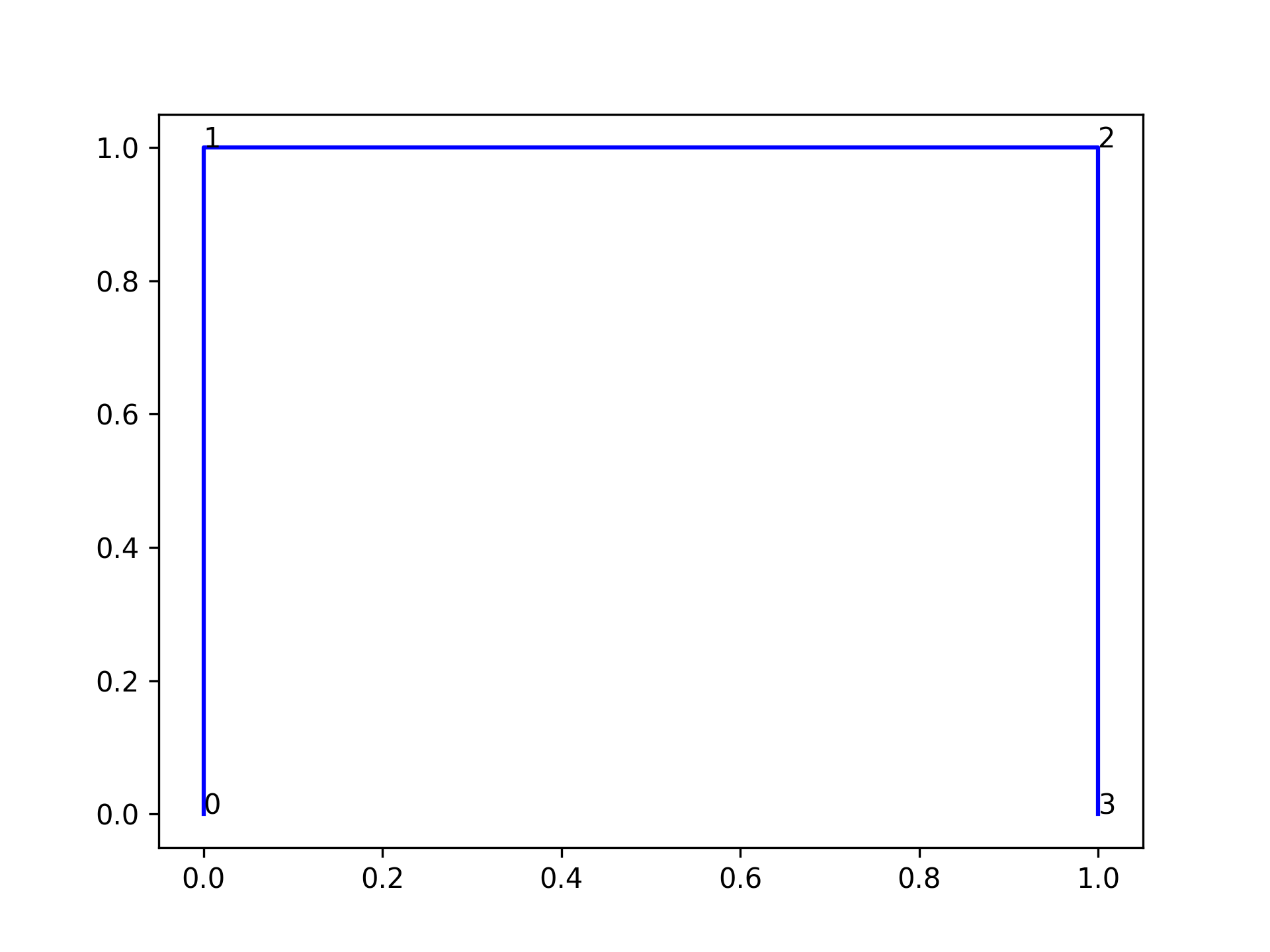}
        \caption{$p=1$}
        \label{hc:p1}
    \end{subfigure}
    \begin{subfigure}{0.32\textwidth}
        \includegraphics[width=\linewidth]{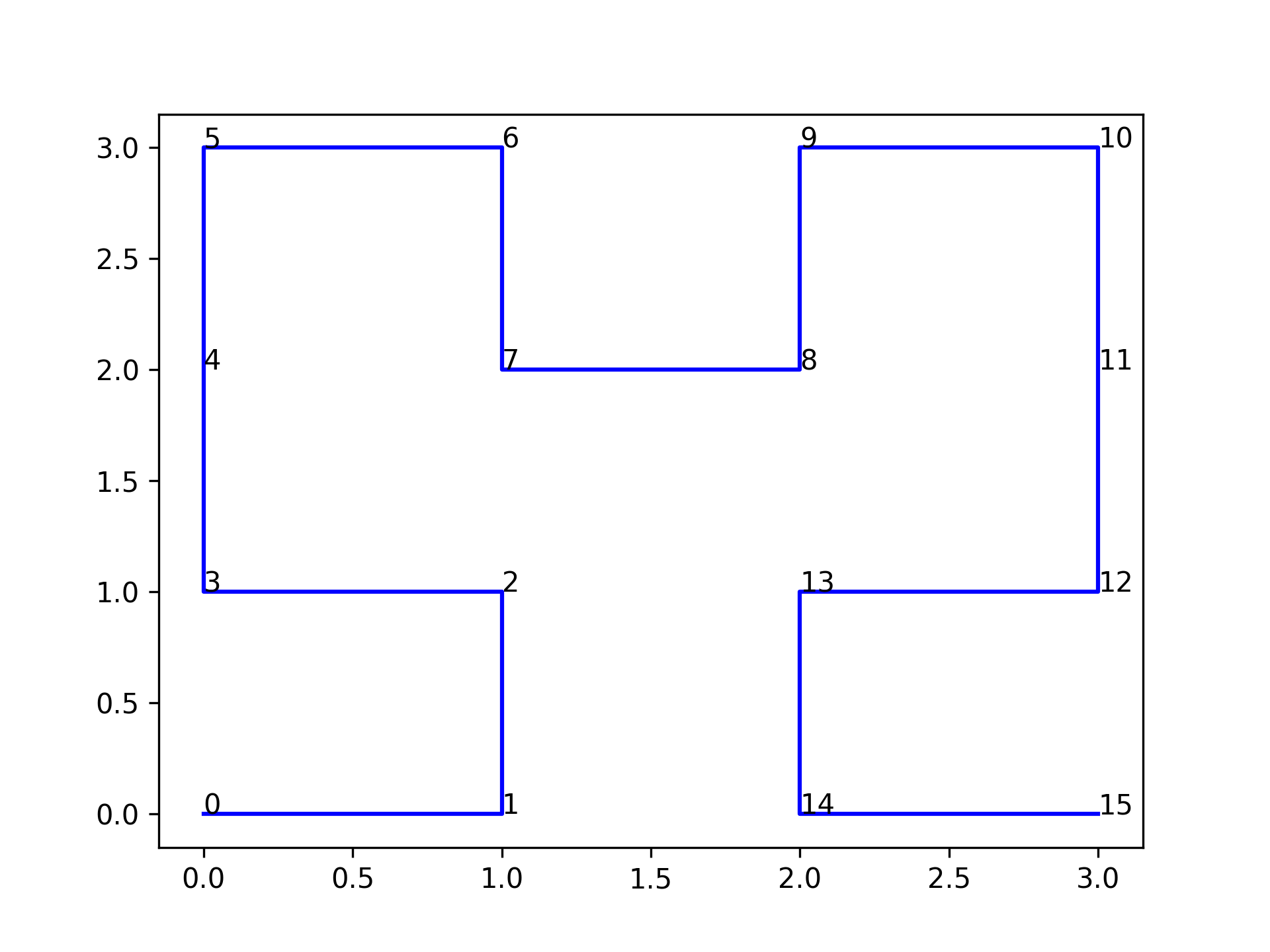}
        \caption{$p=2$}
        \label{hc:p2}
    \end{subfigure}
    \begin{subfigure}{0.32\textwidth}
        \includegraphics[width=\linewidth]{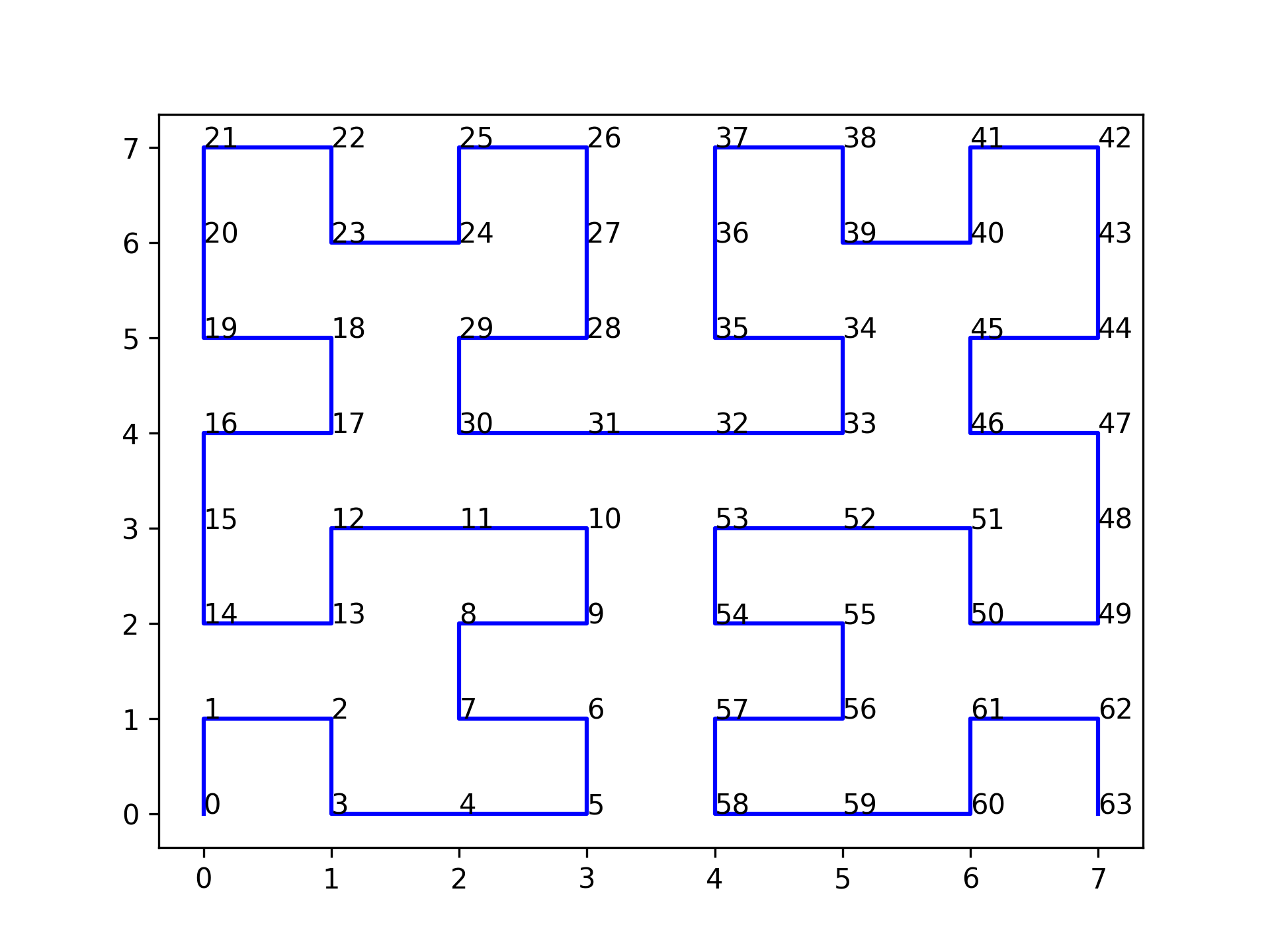}
        \caption{$p=3$}
        \label{hc:p3}
    \end{subfigure}
    \caption{Visualisation of a $2D$ Hilbert curve of order $p = \{1,2,3\}$.}
    \label{fig:hilbert_curve_iterations}
\end{figure}

Consider the construction of a Hilbert curve in $2D$. For the first iteration, a line on the closed interval $[0,1]$ and a square $[0,1]^2$ are taken. Four equidistant points are selected on the line, where the starting point is at $0$ and the end point is at $1$. The square is subdivided into four equal parts. Intervals of the line connecting each pair of points are then mapped onto the square such that the intervals adjacent on the line share a common edge on the square. This results in a simple U-shape, illustrated in Figure~\ref{fig:hilbert_curve_iterations}(\subref{hc:p1}). At each subsequent iteration, the curve from the previous iteration is divided into four equal parts. Each part is then shrunk by a factor of 1/2, rotated, and repositioned such that the four curves connect at their endpoints in a U-shaped or reverse U-shaped pattern. Figure~\ref{fig:hilbert_curve_iterations} shows $2D$ Hilbert curves of order $1$ to $3$. 

The Hilbert curve is bijective, i.e. the mapping between points on the curve and a $d$-dimensional space is reversible. Bijectivity of space-filling curves has been useful in organising multi-dimensional data storage and retrieval systems, since it allows a linear index to the data to be constructed and searched~\cite{Abel1990,Rivest1976}. Ideally, closely related data in the $d$-dimensional space would also be close together on the $1$-dimensional curve. Faloutsos and Roseman~\cite{Faloutsos1989} investigated the clustering properties of Hilbert curves compared to other space-filling curves, and found Hilbert curves to be superior when measuring the average distance (along the curve) to the nearest neighbours of a given point. This finding suggests that Hilbert curves may serve as a viable alternative to nearest neighbour sorting of a sample. 

For the Hilbert curve implementation in this study, we use the \texttt{hilbertcurve} Python package (\url{https://pypi.org/project/hilbertcurve/}), based on the algorithm presented by Skilling \cite{Skilling2004}. All code used for the experiments conducted in this study can be found at \url{https://github.com/jpienaar-tuks/MIT807}.

\section{Hilbert curves as samplers}\label{section:HC_samplers}
Given a Hilbert curve of a particular order, the vertices can be used as a basis for a sample in the associated multidimensional space. Table~\ref{tab:hilbert_function_evals} shows how the number of vertices grows with the order of the curve and the dimension of the search space. When sampling for ELA, it is common practice to use sample sizes of $10^2\times d$ to $10^3\times d$~\cite{Munoz2015_ELA}. 
\begin{table}[!h]
\caption{Exponential growth in Hilbert curve vertices with dimension and curve order. Values that exceed a value of $10^3 \times d$ are formatted in bold.}
\label{tab:hilbert_function_evals}
\setlength{\tabcolsep}{5pt}
\renewcommand*{\arraystretch}{1.15}
\centering
\begin{tabular}{|c|l|r|r|r|r|r|}
\cline{3-7}
\multicolumn{2}{l|}{\multirow{2}{*}{}} & \multicolumn{5}{c|}{Dimension}
\\ \cline{3-7} 
\multicolumn{2}{l|}{}  & 2    & 3    & 5   & 10  & 20    \\ \hline
\multirow{6}{*}{Order}  & 1  & 4  & 8  & 32  & 1024    & $\mathbf{1.05\times 10^6}$  \\ \cline{2-7} 
& 2  & 16   & 64   & 1\,024  & $\mathbf{1.05 \times 10^6}$  & $\mathbf{1.10\times 10^{12}}$ \\ \cline{2-7} 
& 3  & 64 & 512  & \textbf{32\,768}  & $\mathbf{1.07 \times 10^9}$  & $\mathbf{1.15 \times 10^{18}}$ \\ \cline{2-7} 
& 4  & 256  & \textbf{4\,096}  & $\mathbf{1.05 \times 10^6}$  & $\mathbf{1.10 \times 10^{12}}$ & $\mathbf{1.21 \times 10^{24}}$ \\ \cline{2-7} 
& 5  & 1\,024  & \textbf{32\,768}  & $\mathbf{3.36 \times 10^7}$   & $\mathbf{1.13 \times 10^{15}}$  & $\mathbf{1.27 \times 10^{30}}$ \\ \cline{2-7} 
& 6  & \textbf{4\,096} & \textbf{262\,144} & $\mathbf{1.07 \times 10^9}$ & $\mathbf{1.15 \times 10^{18}}$ & $\mathbf{1.33 \times 10^{36}}$ \\ \hline\hline
\multicolumn{1}{|l}{Budget ($10^3 \times d$)}     &      & 2\,000          & 3\,000            & 5\,000                & 10\,000                & 20\,000                \\ \hline
\end{tabular}
\end{table}
Table~\ref{tab:hilbert_function_evals} highlights in bold the order and dimension combinations where the number of vertices exceeds a common sample budget of $10^3 \times d$. To remain within a sampling budget, we sub-sample randomly from the set of vertices on a Hilbert curve with a minimum order of 3. 

\subsection{Stochastic sampling using Hilbert curves}
\label{sec:hc_sampler:stochasticity}

\begin{figure}[!tb]
    \begin{subfigure}{0.5\textwidth}
        \includegraphics[width=\linewidth]{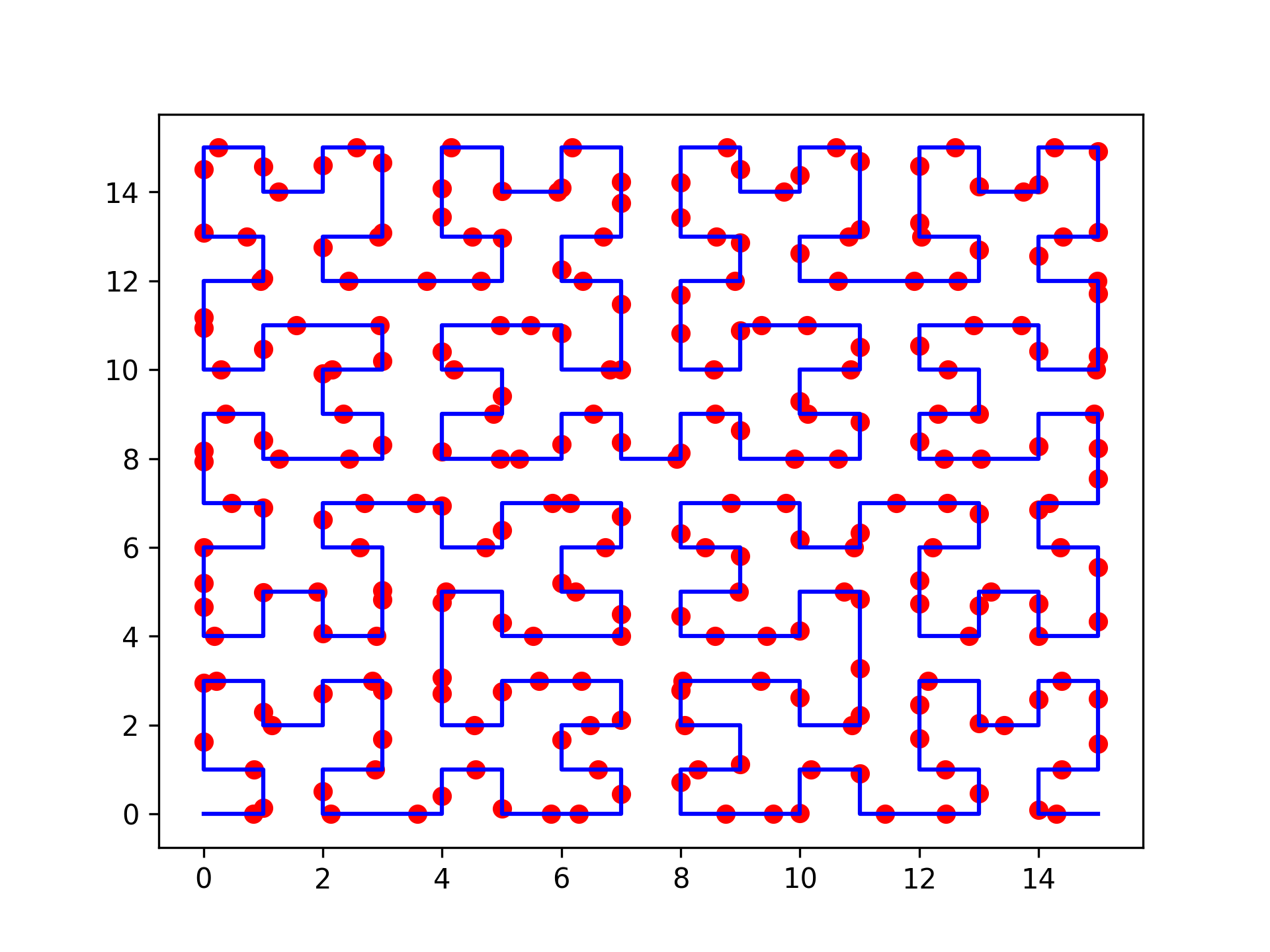}
        \caption{Randomising points along the edges}
        \label{fig:hilbert_along_edges}
    \end{subfigure}
    \begin{subfigure}{0.5\textwidth}
        \includegraphics[width=\linewidth]{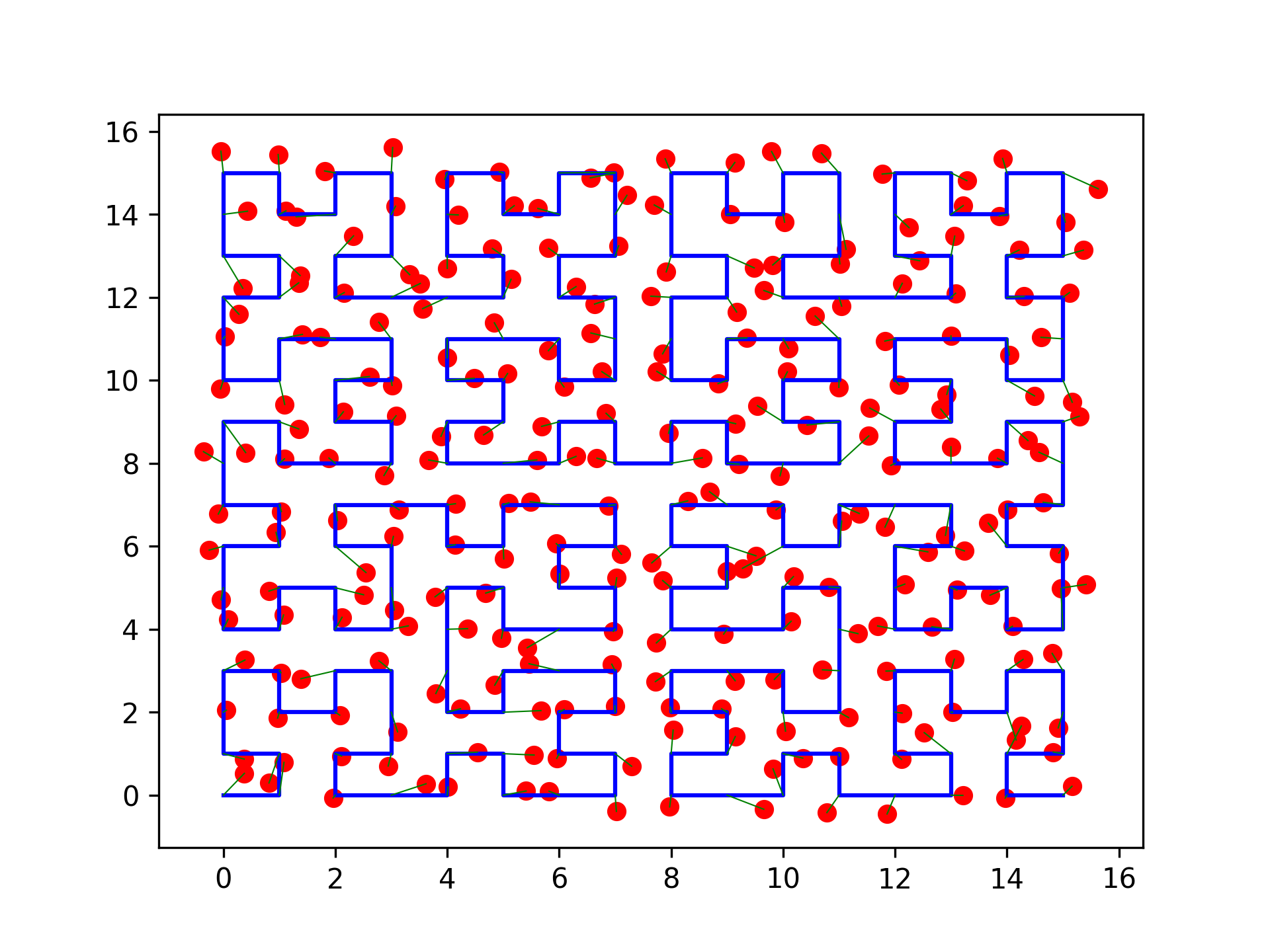}
        \caption{Randomising points near vertices}
        \label{fig:hilbert_near_vertexes}
    \end{subfigure}
    \caption{Illustration in 2D of adding stochasticity to a Hilbert curve}
    \label{fig:hc_stochasticity}
\end{figure}

Since the generation of a Hilbert curve is deterministic, the following two strategies were evaluated to introduce stochasticity:
\begin{itemize}
    \item Selecting random points along the edges of the Hilbert curve, illustrated in Figure~\ref{fig:hc_stochasticity}(\subref{fig:hilbert_along_edges}): Given any two sequential vertices, $P_i$ and $P_{i+1}$, a new point is selected $P_j = rP_i+(1-r)P_{i+1}$, where $r \sim U(0,1)$. 
    \item Selecting points near vertices of the curve, illustrated in Figure~\ref{fig:hc_stochasticity}(\subref{fig:hilbert_near_vertexes}): For each vertex $P_i$, generate a new point drawn from a normal distribution centred on $P_i$ with a standard deviation $\sigma$ (constrained by the bounds of the search space). The step size of the Hilbert curve before scaling is 1, so $\sigma$ was empirically chosen to be 0.3 to prevent excessive potential overlap with points generated by neighbouring vertices.
\end{itemize}

\begin{figure}[!tb]
    \centering
    \includegraphics[width=0.55\textwidth]{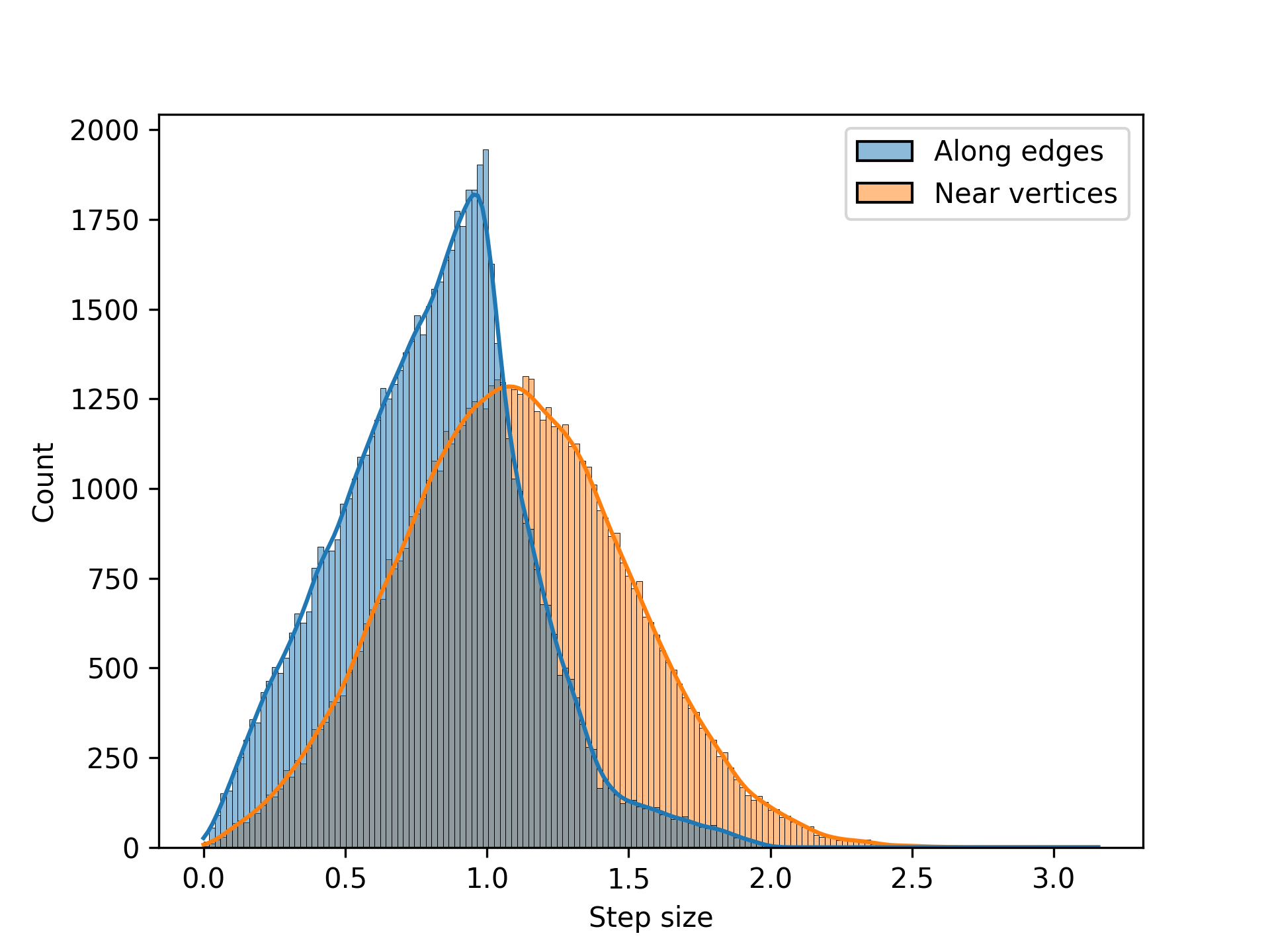}
    \caption{Effect of randomisation strategy on step size. Sample was generated from an $8^{th}$ order Hilbert curve in 2D space resulting in a total of 65\,535 points. The $x$-axis represents the distance between two consecutive points.}
    \label{fig:stochasticity_effect_on_step_size}
\end{figure}

Neighbouring vertices on a Hilbert curve are equidistant, so the sequence of vertices has a constant step size. The two randomisation techniques introduce variation into the step size. 
Figure~\ref{fig:stochasticity_effect_on_step_size} shows that 
the randomisation near vertices strategy produces step sizes with a more normal distribution. For all further experiments we have used the randomisation around vertices strategy.

\subsection{Search space coverage}
\label{sec:hc_sampler:space_coverage}
We now investigate the extent to which Hilbert curve sampling covers the search space compared to competing strategies such as Latin hypercube sampling (LHS). To investigate the search space coverage, the following three strategies were used to draw samples from search space $[-5,5]^d$ of sizes $n \in \{100d, 316d, 1000d\}$ for dimensions $d \in \{5, 10, 20, 30\}$:
\begin{enumerate}
    \item Hilbert curve: randomly select the required number of points from the curve (without replacement).
    \item LHS: generate the required sample using a Latin hypercube design~\cite{McKay2000}.
    \item Random walk: use a simple random walk~\cite{Malan2014} with maximum step size of 1 to generate the sample.
\end{enumerate}
For each sample size and dimension combination, a random uniform sample was drawn as a reference set. Thirty independent runs of each sampling strategy were implemented, and for each run, the Hausdorff distance~\cite{Heinonen2001} to the reference set was calculated. The Hausdorff distance as a measure of search space coverage was first proposed and investigated by Lang and Engelbrecht~\cite{Lang2020a}, and their methodology is followed in this study. Statistical significance tests were performed for each sampling strategy as proposed by Derrac et al.~\cite{Derrac2011} and described by Lang and Engelbrecht~\cite{Lang2020a}.

Table~\ref{tab:hc_sampler:hausdorff} shows that the Hilbert curve and LHS have similar Hausdorff distances, implying a similar coverage of the search space, whereas the random walk provides the worst coverage. Lower values were achieved in most cases by the Hilbert curve sampler, confirmed by the statistical significance tests presented in Table~\ref{tab:hc_sampler:significance}. The null hypothesis is that there is no significant difference between the Hausdorff distances for each of the sampling strategies. Results show that the null hypothesis can be rejected, given the $p$-values and a significance level of 0.05. We therefore conclude that the Hilbert curve sampler achieves a more uniform search space coverage than LHS.

    \begin{table}[t]
    \centering
    \caption{Average Hausdorff distances between samplers and a uniform random sample for different sample sizes and dimensions $d$. Best and worst distances are highlighted in cyan and red, respectively.}
    \label{tab:hc_sampler:hausdorff}
        \begin{tabular}{|l|l|l|l|l|l|} 
            \hline
            \textbf{$D$} & \multicolumn{2}{l|}{\textbf{Sample size }} & \textbf{Hilbert curve}                    & \textbf{Latin hypercube}                     & \textbf{Random walk}                \\ 
            \hline
            5                   & $100\times d$  & 500                                & {\cellcolor{cyan}}\textbf{2.0642 ($\pm$0.0300)}  & 2.0866 ($\pm$0.0242)                             & {\cellcolor{red}}3.9168 ($\pm$0.2967)   \\ 
            \hline
            5                   & $316\times d$  & 1580                               & {\cellcolor{cyan}}\textbf{1.6065 ($\pm$0.0134)}  & 1.6218 ($\pm$0.0121)                             & {\cellcolor{red}}3.9127 ($\pm$0.2731)   \\ 
            \hline
            5                   & $1000\times d$ & 5000                               & 1.2754 ($\pm$0.0097)                              & {\cellcolor{cyan}}\textbf{1.2690 ($\pm$0.0044)} & {\cellcolor{red}}3.8658 ($\pm$0.2952)   \\ 
            \hline
            10                  & $100\times d$  & 1000                               & {\cellcolor{cyan}}\textbf{4.9716 ($\pm$0.0295)}  & 5.0651 ($\pm$0.0300)                             & {\cellcolor{red}}7.6117 ($\pm$0.3323)   \\ 
            \hline
            10                  & $316\times d$  & 3160                               & {\cellcolor{cyan}}\textbf{4.3345 ($\pm$0.0108)}  & 4.4297 ($\pm$0.0143)                             & {\cellcolor{red}}7.5796 ($\pm$0.3217)   \\ 
            \hline
            10                  & $1000\times d$ & 10000                              & {\cellcolor{cyan}}\textbf{3.8045 ($\pm$0.0064)}  & 3.8781 ($\pm$0.0055)                             & {\cellcolor{red}}7.5310 ($\pm$0.3035)   \\ 
            \hline
            20                  & $100\times d$  & 2000                               & {\cellcolor{cyan}}\textbf{9.6778 ($\pm$0.0265)}  & 9.9578 ($\pm$0.0254)                             & {\cellcolor{red}}12.5364 ($\pm$0.2787)  \\ 
            \hline
            20                  & $316\times d$  & 6320                               & {\cellcolor{cyan}}\textbf{8.9312 ($\pm$0.0121)}  & 9.2413 ($\pm$0.0087)                             & {\cellcolor{red}}12.5213 ($\pm$0.2007)  \\ 
            \hline
            20                  & $1000\times d$ & 20000                              & {\cellcolor{cyan}}\textbf{8.2919 ($\pm$0.0082)}  & 8.5967 ($\pm$0.0051)                             & {\cellcolor{red}}12.5013 ($\pm$0.2640)  \\ 
            \hline
            30                  & $100\times d$  & 3000                               & {\cellcolor{cyan}}\textbf{13.5314 ($\pm$0.0263)} & 13.8606 ($\pm$0.0205)                            & {\cellcolor{red}}16.3195 ($\pm$0.2825)  \\ 
            \hline
            30                  & $316\times d$  & 9480                               & {\cellcolor{cyan}}\textbf{12.7478 ($\pm$0.0177)} & 13.1485 ($\pm$0.0065)                            & {\cellcolor{red}}16.2629 ($\pm$0.2186)  \\ 
            \hline
            30                  & $1000\times d$ & 30000                              & {\cellcolor{cyan}}\textbf{12.0615 ($\pm$0.0123)} & 12.4913 ($\pm$0.0057)                            & {\cellcolor{red}}16.3055 ($\pm$0.2677)  \\
            \hline
        \end{tabular}
    \end{table}
    
\begin{table}[!t]
    \centering
    \caption{Ranks achieved by each of the sampling strategies with the $p$-value of the significance tests. Best (lowest) values highlighted in bold.}
    \label{tab:hc_sampler:significance}
    \begin{tabular}{|l|r|r|r|}
        \hline
        Sampler              & Friedman      & Friedman-aligned & Quade          \\
        \hline
        Hilbert curve        & \textbf{1.09} & \textbf{289.23}  & \textbf{1.04}  \\
        \hline
        Latin
          Hypercube    & 1.91          & 431.78           & 1.96           \\
        \hline
        Random walk          & 3.00          & 900.50           & 3.00           \\
        \hline
        $p$-value              & $<$1e-5       & $<$1e-5          & $<$1e-5        \\
        \hline
    \end{tabular}
\end{table}

\subsection{Computational cost}
We now compare the cost of the Hilbert curve as a sampler to alternative sampling strategies. 
Generating the Hilbert curve is not computationally cheap, but we expect that this upfront investment will be justified if the sample is subsequently used to calculate landscape metrics that require ordered samples. 

To evaluate the computational cost, a performance counter\footnote{Python's \texttt{time.perf\_counter}} was used to keep track of the time to a) generate the sample and b) calculate information content metrics for both the Hilbert curve and LHS. Using the \texttt{pflacco} library, the information content (\texttt{ic}) metrics were calculated for each of the 24 BBOB functions defined as part of the COCO platform~\cite{Hansen2020}. This was done for each dimension and sample size listed in Table~\ref{tab:hc_sampler:hausdorff}. Note that for LHS to be used for information content metrics, the sample needs to be ordered. We evaluated both the random ordering and the nearest neighbour ordering strategies. Also note that the information content function provided by the \texttt{pflacco} library was modified to accept \texttt{none} as an ordering argument to use the pre-ordered Hilbert curve samples.

\begin{figure}[b!]
    \begin{subfigure}{0.45\textwidth}
        \includegraphics[width=\linewidth]{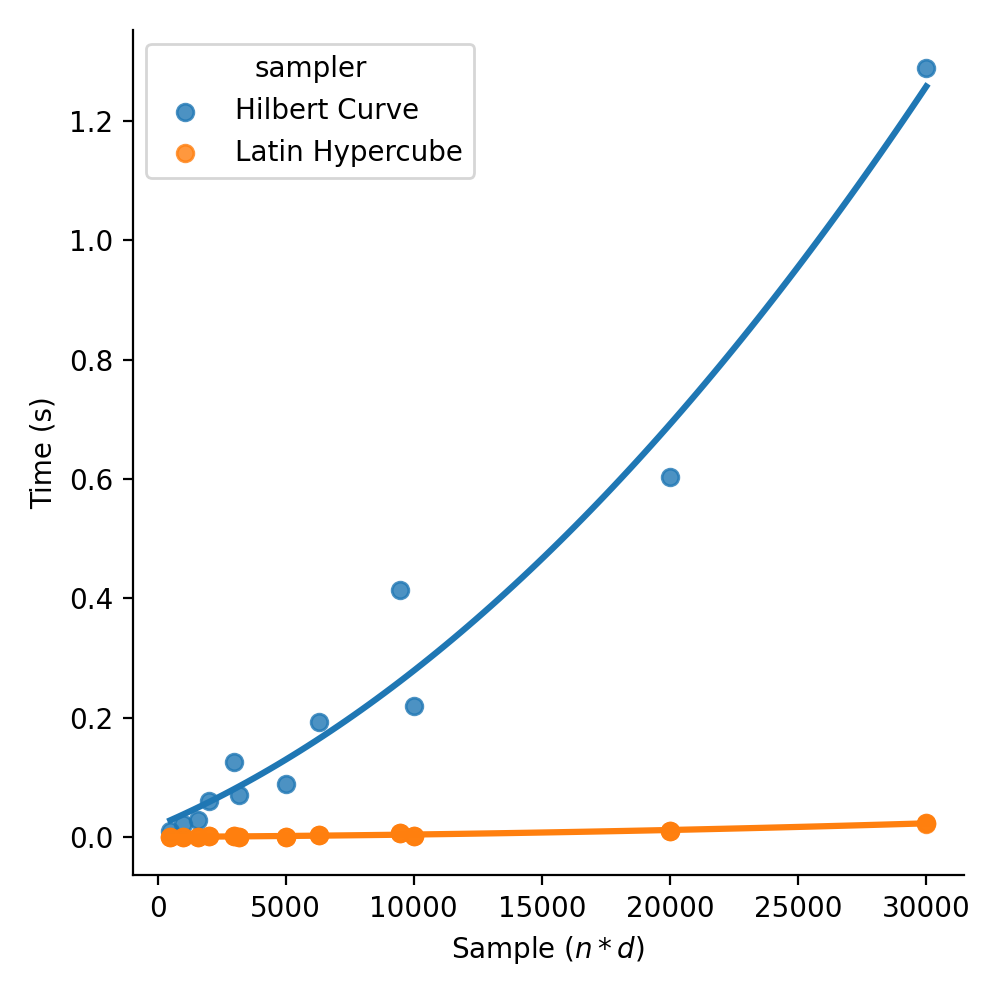}
        \caption{Time required to generate sample only}
        \label{fig:timing:generation}
    \end{subfigure} \quad
    \begin{subfigure}{0.45\textwidth}
        \includegraphics[width=\linewidth]{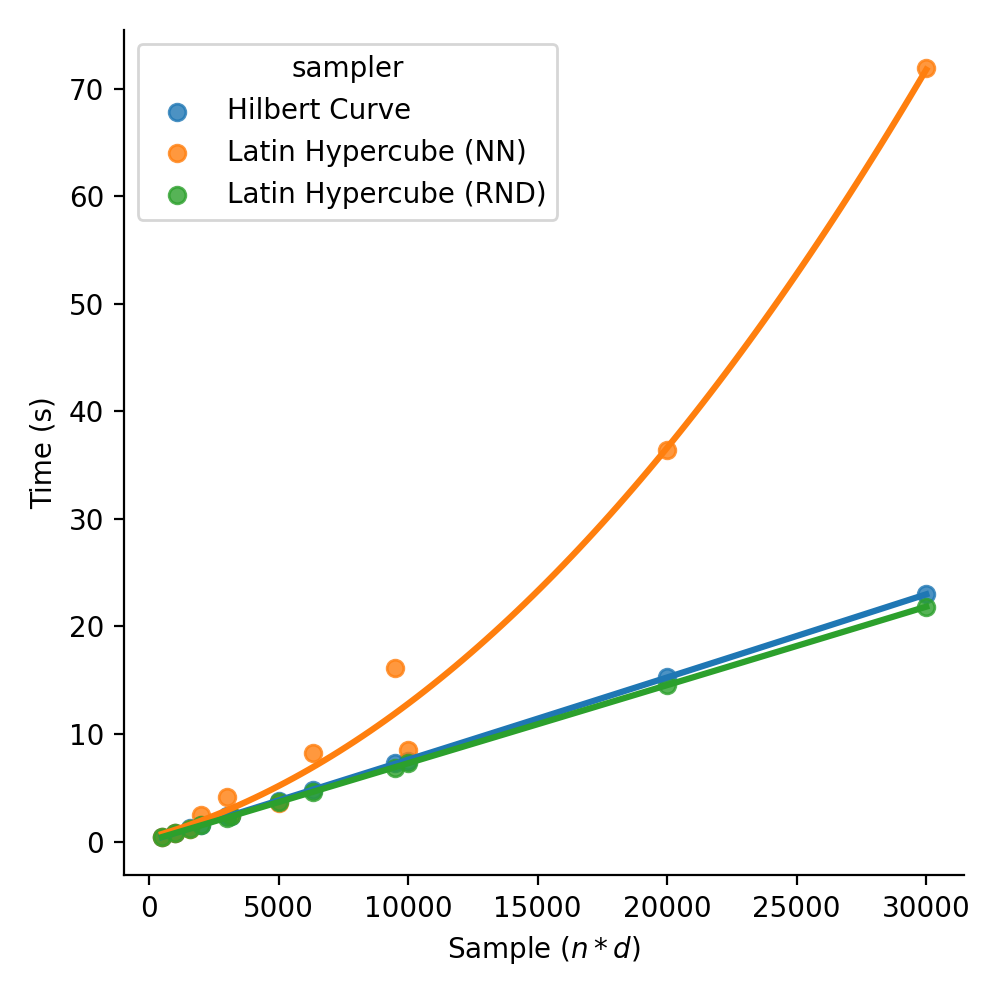}
        \caption{Time to generate sample and calculate information content metrics}
        \label{fig:timing:total}
    \end{subfigure}
    \caption{Comparison of time (in seconds) to generate samples and calculate information content metrics for Hilbert curve sampling and LHS using nearest neighbour and random ordering. Trendlines are polynomials of order 2.}
    \label{fig:hc_sampler:timings}
\end{figure}

Figure~\ref{fig:hc_sampler:timings}(\subref{fig:timing:generation}) 
shows that the Hilbert curve sampler is slower than LHS to generate a sample of the same size. While the computational cost of the Hilbert curve sampler seems to grow at least quadratically, it remains acceptable even for large sample sizes (just over 1.2 seconds to generate a sample of 30\,000 points). 
Figure~\ref{fig:hc_sampler:timings}(\subref{fig:timing:total}) 
shows that for the information content metrics, Hilbert curve sampling strategy is substantially faster (especially at larger sample sizes) than LHS with nearest neighbour ordering.

\subsection{Predictive performance of Hilbert curve samples}
\label{sec:hc_sampler:performance}

We have shown that Hilbert curve sampling provides an efficient method for generating spatially correlated sequences of solutions from continuous search spaces, and that these sequences provide good coverage of the search space. We now investigate whether the landscape features extracted from these samples provide good predictive performance for algorithm selection. 

Following the approach by Mu{\~n}oz et al.~\cite{Munoz2015_ELA}, we use the task of predicting the class of each BBOB function as a proxy for algorithm selection: if the landscape features of function instances can be used to discriminate between problem classes, then they should also be effective predictor variables for algorithm selection. Five target classes corresponding to the BBOB function groupings as outlined in~\cite{hansen2009_BBOB} were used, namely (1) separable functions: $\{f_1\ldots f_5\}$, (2) functions with low or moderate conditioning: $\{f_6\ldots f_9\}$, (3) unimodal functions and functions with high conditioning: $\{f_{10}\ldots f_{14}\}$, (4) multimodal functions with adequate global structure: $\{f_{15}\ldots f_{19}\}$, (5) functions with low or moderate conditioning: $\{f_{20}\ldots f_{24}\}$.

The experimental setup is as follows: 24 BBOB functions in dimensions $d \in \{2,5,10,20\}$ are used, with an evaluation budget of $1\,000 \times d$. Each BBOB function has 15 predefined instances, and three random instances per function are held out as testing data\footnote{This leave-one-instance out approach is used instead of the leave-one-problem out approach due to the small number of classes.}. For each combination of instance and dimension, 1\,000$d$ samples are drawn from $[-5,5]^d$, using three sampling strategies: Hilbert curve (HC), Latin hypercube (LH) and random walk (RW). Four feature sets from the \texttt{pflacco} package were selected that did not require further function evaluations, namely: dispersion (\texttt{disp}), ELA y-distribution (\texttt{ela\_distr}), ELA meta model (\texttt{ela\_meta}), and information content~(\texttt{ic}).\looseness=-1

Table~\ref{tab:accuracy_samplers} gives the testing accuracy of a decision tree, k-nearest neighbour, and random forest classifier 
for the task of predicting the function class from the landscape metrics. The \texttt{scikit-learn} version 1.1.3 default settings were used in all instances. Results show that a random forest model was the most effective at predicting the function class across all sampling strategies. Overall, the RW strategy emerges as the least competitive performer, while the Hilbert curve and Latin hypercube sampling strategies were comparable for all three classifiers.  

    \begin{table}[!tb]
    \centering
    \caption{Accuracy at predicting the function class using four feature sets based on samples from the three different sampling strategies}
    \label{tab:accuracy_samplers}
        \begin{tabular}{|l|c|c|c|} \hline
        & \textbf{Decision Tree} & \textbf{k-Nearest Neighbour} & {\textbf{Random Forest}} \\
        \hline
        { \textbf{HC}} &{ \textbf{$93.50\% (\pm 5.62\%)$}} & { \textbf{$70.15\% (\pm 7.41\%)$}} & { \textbf{$97.38\% (\pm 1.64\%)$}}\\
        \hline
        { \textbf{LH}} &{ \textbf{$93.87\% (\pm 5.01\%)$}} & { \textbf{$72.31\% (\pm 7.34\%)$}} & { {$97.64\% (\pm 2.36\%)$}}\\
        \hline
        { \textbf{RW}}&{ {$88.78\% (\pm 7.06\%)$}} & { {$63.97\% (\pm 6.87\%)$}} & { {$93.60\% (\pm 3.93\%)$}}\\
        \hline
        \end{tabular}
    \end{table}

Table~\ref{tab:hc_sampler:performance} gives the performance for each of the three sampling strategies based on the four feature sets separately. An observable trend is that both the \texttt{ela\_meta} and \texttt{ic} features are very salient features and allow the classifiers to accurately discriminate between the function classes, confirming results of Renau et al.~\cite{Renau2019}. 

    \begin{table}[t]
    \centering
    \caption{Testing accuracy for predicting the function class for different classifiers and different sampling strategies: Hilbert curve (HC), Latin hypercube (LH), random walk (RW), under different dimensions ($D$).}
    \label{tab:hc_sampler:performance}        
        \begin{tabular}{|ll|lll|lll|lll|}
\hline
\multicolumn{2}{|l|}{}                         & \multicolumn{3}{l|}{\textbf{Decision tree}}                                                                                   & \multicolumn{3}{l|}{\textbf{k-Nearest Neighbour}}                                                                             & \multicolumn{3}{l|}{\textbf{Random Forest}}                                                                                   \\ \hline
\multicolumn{1}{|l|}{\textbf{Features}} & \textbf{D} & \multicolumn{1}{l|}{\textbf{HC}} & \multicolumn{1}{l|}{\textbf{LH}} & \textbf{RW}                     & \multicolumn{1}{l|}{\textbf{HC}} & \multicolumn{1}{l|}{\textbf{LH}} & \textbf{RW}                     & \multicolumn{1}{l|}{\textbf{HC}} & \multicolumn{1}{l|}{\textbf{LH}} & \textbf{RW}                     \\ \hline
\multicolumn{1}{|l|}{}                                             & 2          & \multicolumn{1}{l|}{\cellcolor[HTML]{FBE8EB}66.67\%}     & \multicolumn{1}{l|}{\cellcolor[HTML]{FAC1C4}52.78\%}     & \cellcolor[HTML]{F99395}36.11\% & \multicolumn{1}{l|}{\cellcolor[HTML]{FBF4F7}70.83\%}     & \multicolumn{1}{l|}{\cellcolor[HTML]{FBF8FB}72.22\%}     & \cellcolor[HTML]{F98F91}34.72\% & \multicolumn{1}{l|}{\cellcolor[HTML]{FBF8FB}72.22\%}     & \multicolumn{1}{l|}{\cellcolor[HTML]{FBF0F3}69.44\%}     & \cellcolor[HTML]{F99FA1}40.28\% \\ \cline{2-11} 
\multicolumn{1}{|l|}{}                                             & 5          & \multicolumn{1}{l|}{\cellcolor[HTML]{FBD9DB}61.11\%}     & \multicolumn{1}{l|}{\cellcolor[HTML]{FAC9CC}55.56\%}     & \cellcolor[HTML]{F99799}37.50\% & \multicolumn{1}{l|}{\cellcolor[HTML]{FBD9DB}61.11\%}     & \multicolumn{1}{l|}{\cellcolor[HTML]{FACDD0}56.94\%}     & \cellcolor[HTML]{F99395}36.11\% & \multicolumn{1}{l|}{\cellcolor[HTML]{FAC5C8}54.17\%}     & \multicolumn{1}{l|}{\cellcolor[HTML]{FBF0F3}69.44\%}     & \cellcolor[HTML]{F87476}25.00\% \\ \cline{2-11} 
\multicolumn{1}{|l|}{}                                             & 10         & \multicolumn{1}{l|}{\cellcolor[HTML]{FAC9CC}55.56\%}     & \multicolumn{1}{l|}{\cellcolor[HTML]{FAC9CC}55.56\%}     & \cellcolor[HTML]{F87476}25.00\% & \multicolumn{1}{l|}{\cellcolor[HTML]{FAC1C4}52.78\%}     & \multicolumn{1}{l|}{\cellcolor[HTML]{FBE4E7}65.28\%}     & \cellcolor[HTML]{F8696B}20.83\% & \multicolumn{1}{l|}{\cellcolor[HTML]{FBE8EB}66.67\%}     & \multicolumn{1}{l|}{\cellcolor[HTML]{FBE0E3}63.89\%}     & \cellcolor[HTML]{F8787A}26.39\% \\ \cline{2-11} 
\multicolumn{1}{|c|}{\multirow{-4}{*}{\textbf{disp}}}              & 20         & \multicolumn{1}{l|}{\cellcolor[HTML]{FAD1D4}58.33\%}     & \multicolumn{1}{l|}{\cellcolor[HTML]{FBE4E7}65.28\%}     & \cellcolor[HTML]{F98F91}34.72\% & \multicolumn{1}{l|}{\cellcolor[HTML]{FBE4E7}65.28\%}     & \multicolumn{1}{l|}{\cellcolor[HTML]{FBE0E3}63.89\%}     & \cellcolor[HTML]{F8878A}31.94\% & \multicolumn{1}{l|}{\cellcolor[HTML]{FBE0E3}63.89\%}     & \multicolumn{1}{l|}{\cellcolor[HTML]{FBE8EB}66.67\%}     & \cellcolor[HTML]{F86C6E}22.22\% \\ \hline
\multicolumn{1}{|l|}{}                                             & 2          & \multicolumn{1}{l|}{\cellcolor[HTML]{FAD1D4}58.33\%}     & \multicolumn{1}{l|}{\cellcolor[HTML]{FAB6B8}48.61\%}     & \cellcolor[HTML]{F98F91}34.72\% & \multicolumn{1}{l|}{\cellcolor[HTML]{FAC1C4}52.78\%}     & \multicolumn{1}{l|}{\cellcolor[HTML]{FAC9CC}55.56\%}     & \cellcolor[HTML]{F88B8E}33.33\% & \multicolumn{1}{l|}{\cellcolor[HTML]{FABEC0}51.39\%}     & \multicolumn{1}{l|}{\cellcolor[HTML]{FABEC0}51.39\%}     & \cellcolor[HTML]{F8878A}31.94\% \\ \cline{2-11} 
\multicolumn{1}{|l|}{}                                             & 5          & \multicolumn{1}{l|}{\cellcolor[HTML]{FBE8EB}66.67\%}     & \multicolumn{1}{l|}{\cellcolor[HTML]{FBE8EB}66.67\%}     & \cellcolor[HTML]{F88B8E}33.33\% & \multicolumn{1}{l|}{\cellcolor[HTML]{FAD1D4}58.33\%}     & \multicolumn{1}{l|}{\cellcolor[HTML]{FBDDDF}62.50\%}     & \cellcolor[HTML]{F88486}30.56\% & \multicolumn{1}{l|}{\cellcolor[HTML]{FBE0E3}63.89\%}     & \multicolumn{1}{l|}{\cellcolor[HTML]{FBECEF}68.06\%}     & \cellcolor[HTML]{F99FA1}40.28\% \\ \cline{2-11} 
\multicolumn{1}{|l|}{}                                             & 10         & \multicolumn{1}{l|}{\cellcolor[HTML]{FCFCFF}73.61\%}     & \multicolumn{1}{l|}{\cellcolor[HTML]{D4E0F1}79.17\%}     & \cellcolor[HTML]{F99FA1}40.28\% & \multicolumn{1}{l|}{\cellcolor[HTML]{FBF4F7}70.83\%}     & \multicolumn{1}{l|}{\cellcolor[HTML]{FBECEF}68.06\%}     & \cellcolor[HTML]{F9AEB1}45.83\% & \multicolumn{1}{l|}{\cellcolor[HTML]{DEE7F5}77.78\%}     & \multicolumn{1}{l|}{\cellcolor[HTML]{F2F5FC}75.00\%}     & \cellcolor[HTML]{F99395}36.11\% \\ \cline{2-11} 
\multicolumn{1}{|c|}{\multirow{-4}{*}{\textbf{ela\_distr}}}        & 20         & \multicolumn{1}{l|}{\cellcolor[HTML]{D4E0F1}79.17\%}     & \multicolumn{1}{l|}{\cellcolor[HTML]{F2F5FC}75.00\%}     & \cellcolor[HTML]{F9A3A5}41.67\% & \multicolumn{1}{l|}{\cellcolor[HTML]{FBF0F3}69.44\%}     & \multicolumn{1}{l|}{\cellcolor[HTML]{FBE0E3}63.89\%}     & \cellcolor[HTML]{F9A6A9}43.06\% & \multicolumn{1}{l|}{\cellcolor[HTML]{E8EEF8}76.39\%}     & \multicolumn{1}{l|}{\cellcolor[HTML]{E8EEF8}76.39\%}     & \cellcolor[HTML]{F9A6A9}43.06\% \\ \hline
\multicolumn{1}{|l|}{}                                             & 2          & \multicolumn{1}{l|}{\cellcolor[HTML]{D4E0F1}79.17\%}     & \multicolumn{1}{l|}{\cellcolor[HTML]{D4E0F1}79.17\%}     & \cellcolor[HTML]{FAD5D8}59.72\% & \multicolumn{1}{l|}{\cellcolor[HTML]{FBDDDF}62.50\%}     & \multicolumn{1}{l|}{\cellcolor[HTML]{FBE8EB}66.67\%}     & \cellcolor[HTML]{FAB6B8}48.61\% & \multicolumn{1}{l|}{\cellcolor[HTML]{A1BCDF}86.11\%}     & \multicolumn{1}{l|}{\cellcolor[HTML]{C0D2EA}81.94\%}     & \cellcolor[HTML]{D4E0F1}79.17\% \\ \cline{2-11} 
\multicolumn{1}{|l|}{}                                             & 5          & \multicolumn{1}{l|}{\cellcolor[HTML]{B6CBE7}83.33\%}     & \multicolumn{1}{l|}{\cellcolor[HTML]{A1BCDF}86.11\%}     & \cellcolor[HTML]{B6CBE7}83.33\% & \multicolumn{1}{l|}{\cellcolor[HTML]{F2F5FC}75.00\%}     & \multicolumn{1}{l|}{\cellcolor[HTML]{D4E0F1}79.17\%}     & \cellcolor[HTML]{FAD5D8}59.72\% & \multicolumn{1}{l|}{\cellcolor[HTML]{97B5DC}87.50\%}     & \multicolumn{1}{l|}{\cellcolor[HTML]{6F99CE}93.06\%}     & \cellcolor[HTML]{ACC4E3}84.72\% \\ \cline{2-11} 
\multicolumn{1}{|l|}{}                                             & 10         & \multicolumn{1}{l|}{\cellcolor[HTML]{5A8AC6}95.83\%}     & \multicolumn{1}{l|}{\cellcolor[HTML]{83A7D5}90.28\%}     & \cellcolor[HTML]{A1BCDF}86.11\% & \multicolumn{1}{l|}{\cellcolor[HTML]{E8EEF8}76.39\%}     & \multicolumn{1}{l|}{\cellcolor[HTML]{97B5DC}87.50\%}     & \cellcolor[HTML]{FAC1C4}52.78\% & \multicolumn{1}{l|}{\cellcolor[HTML]{79A0D1}91.67\%}     & \multicolumn{1}{l|}{\cellcolor[HTML]{6592CA}94.44\%}     & \cellcolor[HTML]{8DAED8}88.89\% \\ \cline{2-11} 
\multicolumn{1}{|c|}{\multirow{-4}{*}{\textbf{ela\_meta}}}         & 20         & \multicolumn{1}{l|}{\cellcolor[HTML]{79A0D1}91.67\%}     & \multicolumn{1}{l|}{\cellcolor[HTML]{8DAED8}88.89\%}     & \cellcolor[HTML]{B6CBE7}83.33\% & \multicolumn{1}{l|}{\cellcolor[HTML]{C0D2EA}81.94\%}     & \multicolumn{1}{l|}{\cellcolor[HTML]{97B5DC}87.50\%}     & \cellcolor[HTML]{FABABC}50.00\% & \multicolumn{1}{l|}{\cellcolor[HTML]{6592CA}94.44\%}     & \multicolumn{1}{l|}{\cellcolor[HTML]{79A0D1}91.67\%}     & \cellcolor[HTML]{97B5DC}87.50\% \\ \hline
\multicolumn{1}{|c|}{}                                             & 2          & \multicolumn{1}{l|}{\cellcolor[HTML]{83A7D5}90.28\%}     & \multicolumn{1}{l|}{\cellcolor[HTML]{6592CA}94.44\%}     & \cellcolor[HTML]{C0D2EA}81.94\% & \multicolumn{1}{l|}{\cellcolor[HTML]{83A7D5}90.28\%}     & \multicolumn{1}{l|}{\cellcolor[HTML]{8DAED8}88.89\%}     & \cellcolor[HTML]{ACC4E3}84.72\% & \multicolumn{1}{l|}{\cellcolor[HTML]{83A7D5}90.28\%}     & \multicolumn{1}{l|}{\cellcolor[HTML]{97B5DC}87.50\%}     & \cellcolor[HTML]{97B5DC}87.50\% \\ \cline{2-11} 
\multicolumn{1}{|c|}{}                                             & 5          & \multicolumn{1}{l|}{\cellcolor[HTML]{83A7D5}90.28\%}     & \multicolumn{1}{l|}{\cellcolor[HTML]{83A7D5}90.28\%}     & \cellcolor[HTML]{ACC4E3}84.72\% & \multicolumn{1}{l|}{\cellcolor[HTML]{ACC4E3}84.72\%}     & \multicolumn{1}{l|}{\cellcolor[HTML]{79A0D1}91.67\%}     & \cellcolor[HTML]{A1BCDF}86.11\% & \multicolumn{1}{l|}{\cellcolor[HTML]{6F99CE}93.06\%}     & \multicolumn{1}{l|}{\cellcolor[HTML]{6F99CE}93.06\%}     & \cellcolor[HTML]{79A0D1}91.67\% \\ \cline{2-11} 
\multicolumn{1}{|c|}{}                                             & 10         & \multicolumn{1}{l|}{\cellcolor[HTML]{97B5DC}87.50\%}     & \multicolumn{1}{l|}{\cellcolor[HTML]{79A0D1}91.67\%}     & \cellcolor[HTML]{79A0D1}91.67\% & \multicolumn{1}{l|}{\cellcolor[HTML]{A1BCDF}86.11\%}     & \multicolumn{1}{l|}{\cellcolor[HTML]{6F99CE}93.06\%}     & \cellcolor[HTML]{A1BCDF}86.11\% & \multicolumn{1}{l|}{\cellcolor[HTML]{79A0D1}91.67\%}     & \multicolumn{1}{l|}{\cellcolor[HTML]{5A8AC6}95.83\%}     & \cellcolor[HTML]{6592CA}94.44\% \\ \cline{2-11} 
\multicolumn{1}{|c|}{\multirow{-4}{*}{\textbf{ic}}}                & 20         & \multicolumn{1}{l|}{\cellcolor[HTML]{8DAED8}88.89\%}     & \multicolumn{1}{l|}{\cellcolor[HTML]{ACC4E3}84.72\%}     & \cellcolor[HTML]{6592CA}94.44\% & \multicolumn{1}{l|}{\cellcolor[HTML]{FCFCFF}73.61\%}     & \multicolumn{1}{l|}{\cellcolor[HTML]{79A0D1}91.67\%}     & \cellcolor[HTML]{8DAED8}88.89\% & \multicolumn{1}{l|}{\cellcolor[HTML]{A1BCDF}86.11\%}     & \multicolumn{1}{l|}{\cellcolor[HTML]{6F99CE}93.06\%}     & \cellcolor[HTML]{6592CA}94.44\% \\ \hline
\end{tabular}
    \end{table}
 

\section{Hilbert curves as an ordering tool}\label{section:HC_as_sorters}
In this section, we investigate the use of Hilbert curves as an ordering aid for samples generated from other sampling methodologies. 
A Hilbert curve of order $p$ maps each point on a 1-dimensional curve $[0,\dots,2^{dp}]$ to a point in $[0,\dots,2^p-1]^d$ and vice versa, as illustrated earlier in Figure~\ref{fig:hilbert_curve_iterations}. The mapping is bijective, and allows for an efficient mapping from a point in the $d$-dimensional space to a point on the 1-dimensional Hilbert curve. 

This $d$-D to 1-D mapping can be exploited, given that a Latin hypercube sample of $n$ points divides each axis into $n$ intervals. By selecting the appropriate curve order $p$, i.e. 
    $
        p=\lceil{log_2(n+1)}\rceil
    $,
we can ensure that the resulting Hilbert curve is of sufficient length to accurately map all LHS points to points on the Hilbert curve. The ordering of the points on the Hilbert curve can then be used to provide a spatially correlated ordering of the Latin hypercube sample.

Given that the current best practice of using a greedy nearest neighbour strategy to order LHS requires calculating at least half of the pairwise distance matrix between all points in the sample, the Hilbert curve ordering strategy has the potential to be significantly faster. Furthermore, the nearest neighbour strategy tends to start with relatively short step sizes, which increase as the number of unvisited points decrease. Hilbert curve ordering is likely to provide more consistent step sizes.

\subsection{Step size consistency}
\begin{figure}[!t]
     \begin{subfigure}{0.48\linewidth}
         \includegraphics[width=\linewidth]{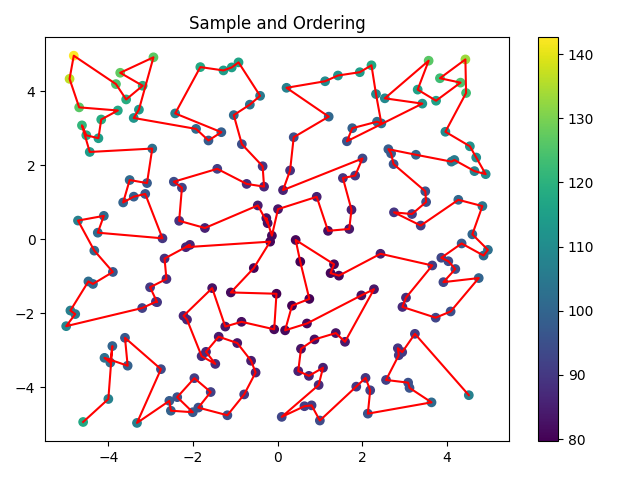}
         \caption{Ordering using Hilbert curves}
         \label{fig:ordering_viz:hc}
     \end{subfigure}
     \begin{subfigure}{0.48\linewidth}
         \includegraphics[width=\linewidth]{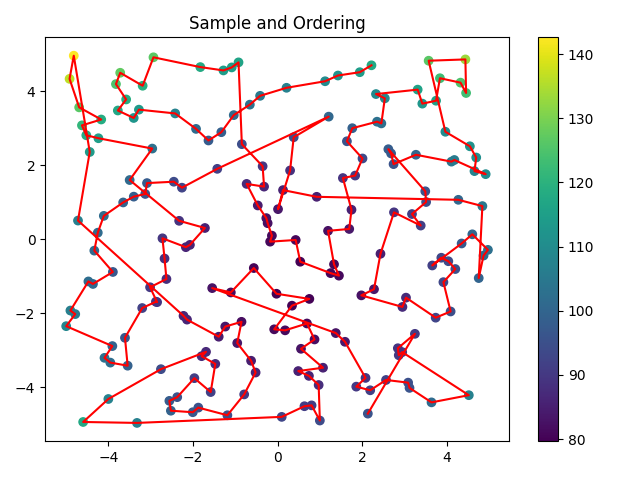}
         \caption{Ordering using nearest neighbour}
         \label{fig:ordering_viz:nn}
     \end{subfigure}
    
    \begin{subfigure}{0.48\linewidth}
         \includegraphics[width=\linewidth]{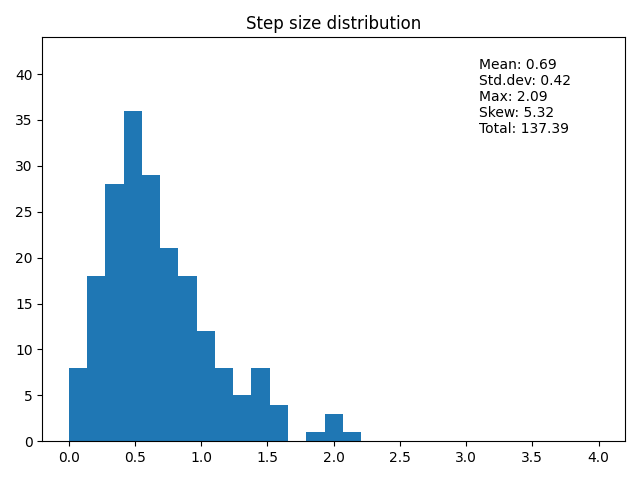}
         \caption{Step sizes  using Hilbert curves}
         \label{fig:step_size_viz:hc}
     \end{subfigure}
     \begin{subfigure}{0.48\linewidth}
         \includegraphics[width=\linewidth]{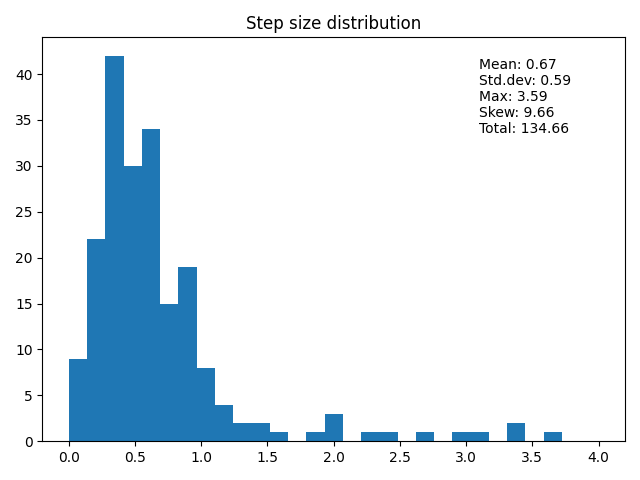}
         \caption{Step sizes  using nearest neighbour}
         \label{fig:step_size_viz:nn}
     \end{subfigure}
    
        \caption{Visualisation of sample ordering using Hilbert curve (left) and nearest neighbour (right) ordering strategies.}
        \label{fig:ordering_viz}
        
\end{figure}

Figure \ref{fig:ordering_viz} shows a visual comparison of the Hilbert curve and nearest neighbour ordering strategies on a $2D$ Sphere function sample. The underlying sample $X$ for both strategies is identical, and was obtained using LHS. In the figures the orderings produced by the Hilbert curve (HC) and the nearest neighbour (NN) approach are depicted as a red line. Colour scale is used to indicate the fitness values of the sampled points. It is evident from Figure \ref{fig:ordering_viz} that the maximum step size is smaller for HC than for NN. The step size distributions shown in Figures \ref{fig:ordering_viz}(\subref{fig:step_size_viz:hc}) and \ref{fig:ordering_viz}(\subref{fig:step_size_viz:nn}) confirm that the maximum step size of HC is lower, and show a greater skew in the step size distribution for NN.


\subsection{Computational cost}
To evaluate the computational cost of the Hilbert curve ordering, we compare it to the nearest neighbour and random (RND) ordering strategies applied to a sample generated using LHS. The Latin hypercube samples of sizes $n \in \{100d, 316d, 1000d\}$ were drawn from search space $[-5,5]^d$ for dimensions $d \in \{5, 10, 20, 30\}$. For each configuration, 24 independent samples were drawn, and timed with a performance counter (Python's \texttt{time.perf\_counter}). 

\begin{figure}[!tb]
    \centering
    \includegraphics[width=0.45\linewidth]{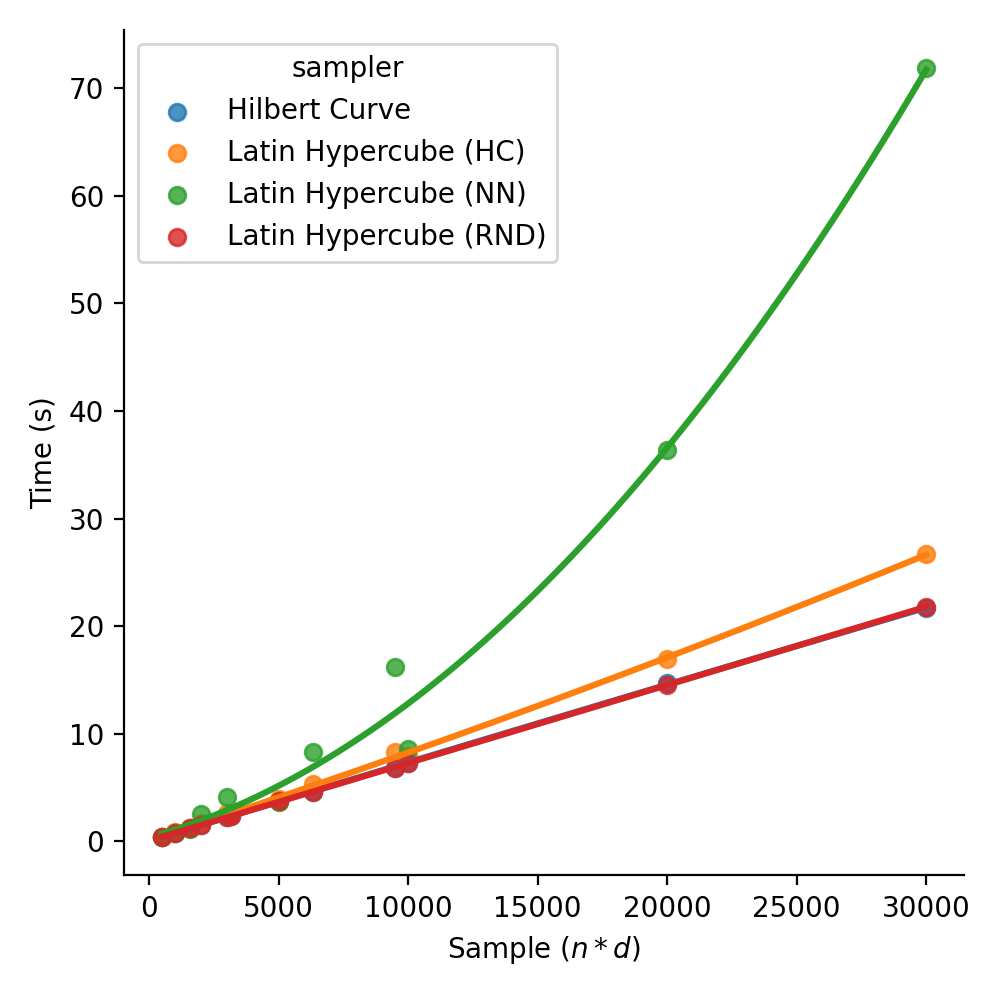}
    \caption{Comparison of time (in seconds) to order the samples generated by the Latin hypercube sampling strategy using Hilbert curve, nearest neighbour, and random ordering. Trendlines are polynomials of order 2.}
    \label{fig:timing:results}
\end{figure}

Figure \ref{fig:timing:results} shows the performance comparison between the various ordering strategies. Hilbert curve sampling as discussed in Section~\ref{section:HC_samplers} is included for completeness, as it does not require additional ordering. From Figure \ref{fig:timing:results}, NN ordering performs the worst, especially for large sample sizes. Furthermore, RND sorting and Hilbert curve \textit{sampling} are equally fast (red and blue curves overlap). Hilbert curve \textit{ordering} produces an intermediate result, slightly slower than random ordering, but significantly faster than the NN ordering. Since the underlying sample is generated by LHS, Hilbert curve ordering retains all the benefits of LHS for ELA features insensitive to order (such as \texttt{ela\_meta}, or \texttt{disp}).

\subsection{Evaluation of features generated using Hilbert curve ordering}
\label{hc_order:eval}
We now compare the information content features generated from a Latin hypercube sample using the HC, NN, and RND ordering. Since the feature values are evaluated rather than the computational cost, this experiment was restricted to dimensions $d\in\{2,5\}$ and sample sizes $n\in\{100d, 1000d\}$. A single sample set $X$ was generated and evaluated on each of the BBOB functions to generate corresponding $y = f(X)$ results. Information content metrics were then calculated on these $(X, y)$ pairs using each of the ordering strategies. This procedure was repeated 30 times.

\begin{figure}[!t]
    \centering
    \includegraphics[width=\linewidth]{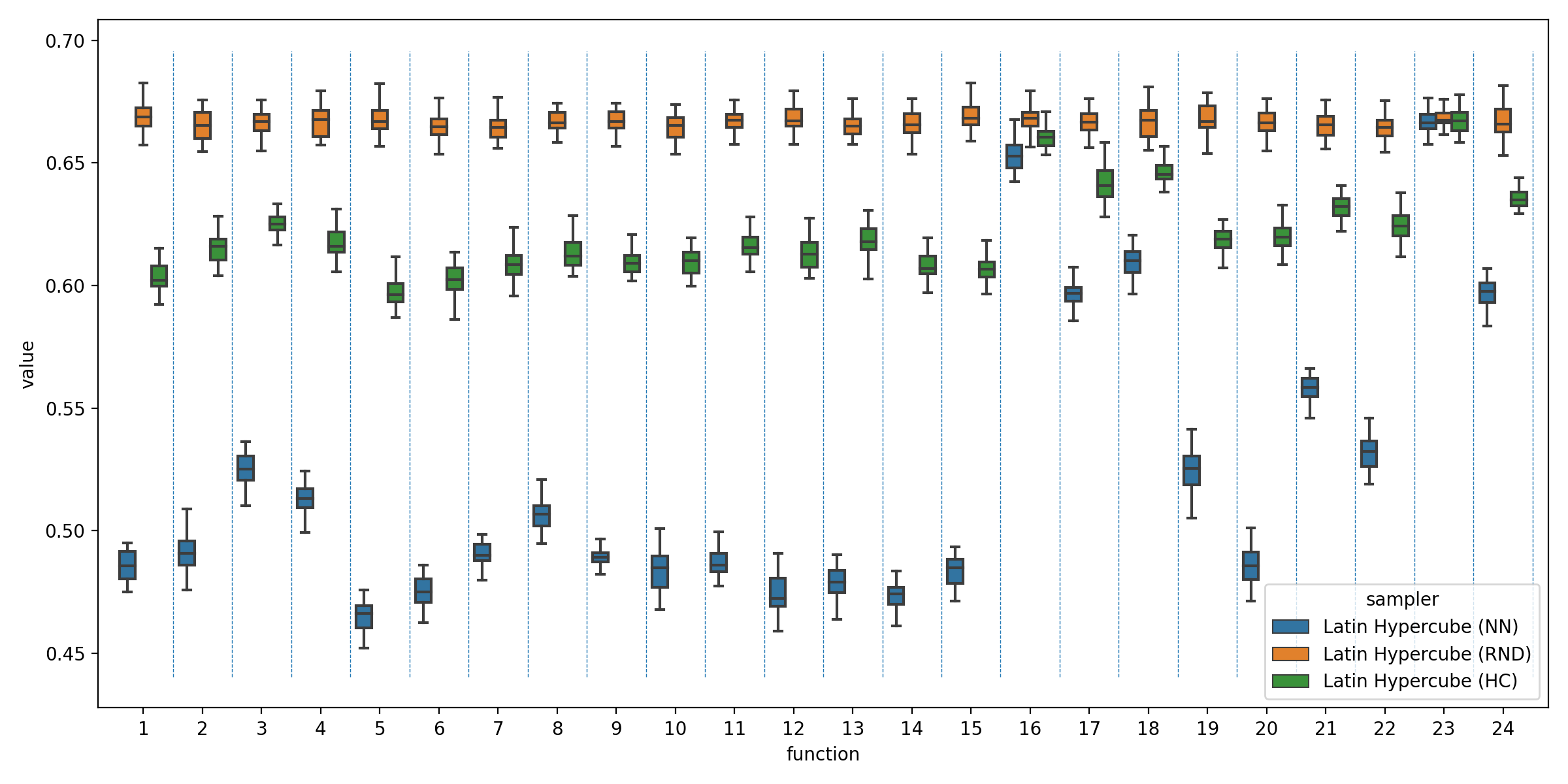}
    \caption{$M_0$ in $5D$ as calculated using nearest neighbour (NN), random (RND) and Hilbert curve (HC) ordering methodologies ($n=1000d$).}
    \label{fig:evals:m0}
\end{figure}

\begin{figure}[!t]
    \centering
    \includegraphics[width=\linewidth]{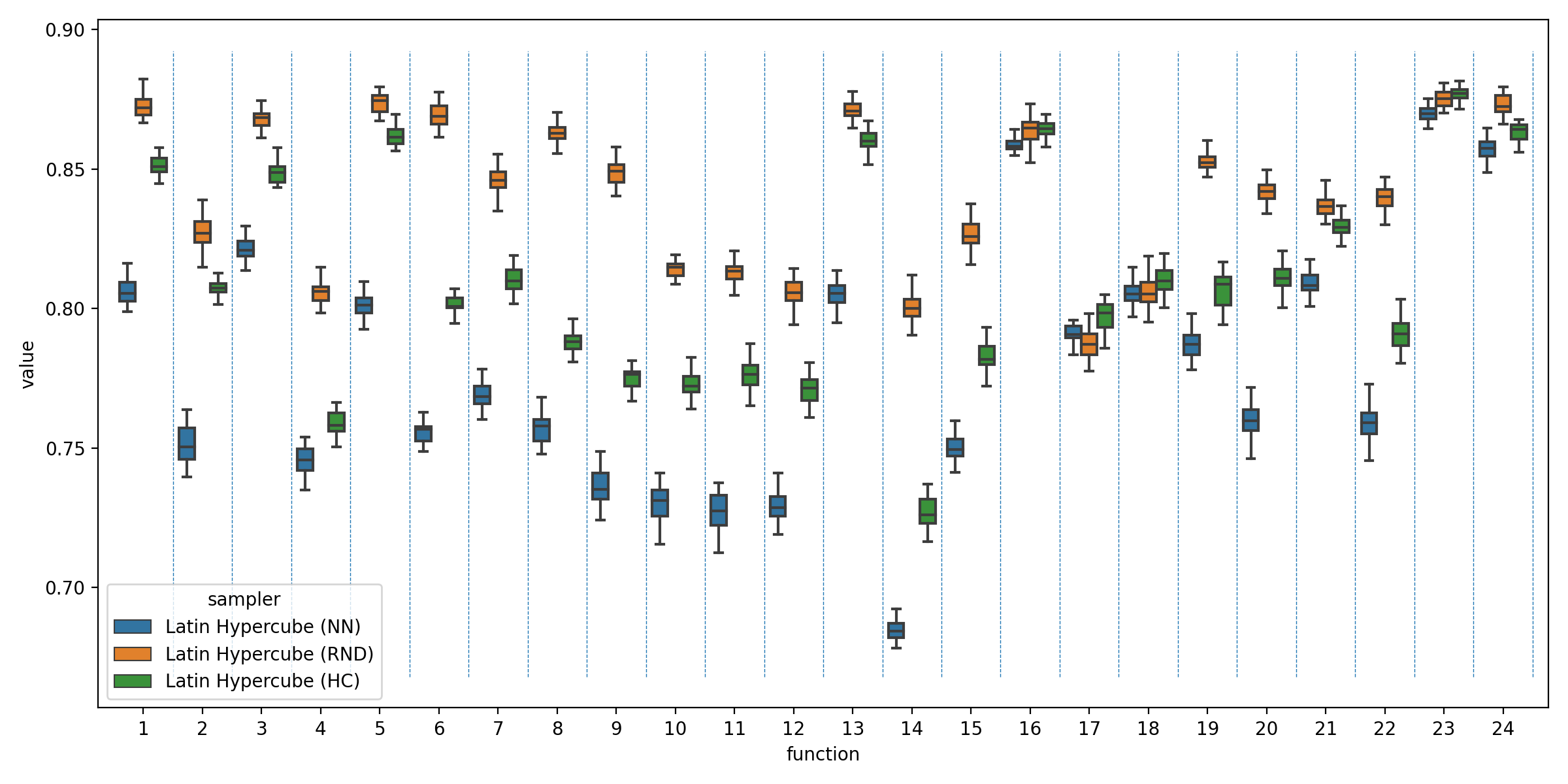}
    \caption{$H_{max}$ in $5D$ as calculated using nearest neighbour (NN), random (RND) and Hilbert curve (HC) ordering methodologies ($n=1000d$).}
    \label{fig:evals:h_max}
\end{figure}

Figures \ref{fig:evals:m0} and \ref{fig:evals:h_max} present initial partial information ($M_0$) and maximum information content ($H_{max}$) landscape features based on the entropy of the sample. We consider BBOB problems in $5D$ (problem indices listed on the $x$-axis), and calculate feature values based on NN, RND, and HC ordering for $n=1000d$.

It is evident from Figure \ref{fig:evals:m0} that the HC ordering produces values that lie between RND and NN, but follow a pattern similar to that of NN. It is clear that $M_0^{RND}$ converges to the same value regardless of the underlying function (confirming the results of ~\cite{Munoz2015_ELA}). $M_0^{NN}$ and $M_0^{HC}$, on the other hand, produce values that allow for discrimination between functions. Since larger values for $M_0$ indicate a more rugged landscape (more changes in concavity over the length of the walk), the HC ordering would on average indicate a more rugged landscape than the NN ordering.

According to Figure~\ref{fig:evals:h_max}, $H_{max}^{RND}$ varies over a wider range than $M_0^{RND}$. However, the $H_{max}^{RND}$ values still tend to cluster around the upper edge of the graph, closer to the maximum value of 1. In comparison, $H_{max}^{NN}$ and $H_{max}^{HC}$ vary over a larger part of the $[0,1]$ range. Larger values for $H_{max}$ indicate a larger variety of objects in the landscape (pits, peaks, plateaus). Thus, HC ordering again indicates a more rugged landscape than the NN ordering.


To evaluate the relative saliency of the features produced by the various ordering strategies, permutation-based feature importance technique~\cite{Dwivedi2023} is employed. We train a random forest (RF) classifier using information content features as input ($\{\epsilon_s, 
\epsilon_{max}, \epsilon_{0.5}, H_{max} \text{ and } M_0\}$, as defined in~\cite{Munoz2015_ELA}), and the function class as output (as defined in~\cite{Munoz2015_ELA} and used in Section~\ref{section:HC_samplers}). Feature saliency is gauged by randomly permuting each input feature in turn across the dataset, and observing the corresponding reduction in RF accuracy. For this purpose, the features are grouped by the ordering strategy, and a RF is trained on $\frac{2}{3}$ of the data, with the remaining $\frac{1}{3}$ held out for testing.  

After training, the base accuracy of the RF classifier is evaluated on the held out data. The accuracy of the RF classifier is then re-evaluated as each feature in the held-out dataset is randomly permuted in turn, and the difference to the base accuracy is recorded. This procedure is repeated 10 times for each sampling strategy. The results are shown in Table \ref{tab:hc_sorter:feature_permutations}. 

Table~\ref{tab:hc_sorter:feature_permutations} shows that  $\epsilon_s$ was the most salient feature across all sample orderings, with the RF trained on the HC ordered features relying on $\epsilon_s$ more heavily than the RF trained on the NN or RND ordered features. RND sample ordering made the $M_0$ feature near useless for the classifier, in line with the observations made in Figure~\ref{fig:evals:m0}. Conversely, RF trained on the NN ordering relied on $M_0$ more than RF trained on the HC ordered features. Overall, HC ordering yielded marginally better RF performance compared to the NN and RND ordering approaches.

\begin{table}[!htb]
\centering
    \caption{Impact of permuting information content features on the accuracy of a random forest (RF) classifier. Classification target was the function groupings.}
       \begin{tabular}{|l|c|c|c|}
            \hline
            Sampler        & LHS (HC)                       & LHS (NN)                       & LHS (RND)                      \\ \hline
            RF base accuracy           & 98.44\% (±0.39\%)                          & 98.25\% (±0.37\%)                          & 97.44\% (±0.54\%)                          \\ \hline \hline
            $\epsilon_{max}$ (\texttt{ic.eps\_max})    & \cellcolor[HTML]{B6CBE7}-8.17\% (±2.22\%)  & \cellcolor[HTML]{BACDE8}-8.46\% (±1.98\%)  & \cellcolor[HTML]{FCFCFF}-14.36\% (±2.76\%) \\ \hline
            $\epsilon_{0.5}$ (\texttt{ic.eps\_ratio}) & \cellcolor[HTML]{FBE3E6}-20.62\% (±3.28\%) & \cellcolor[HTML]{D0DDF0}-10.45\% (±1.64\%) & \cellcolor[HTML]{FAC7CA}-27.82\% (±3.23\%) \\ \hline
            $\epsilon_s$ (\texttt{ic.eps\_s})  & \cellcolor[HTML]{F8696B}-52.40\% (±2.47\%) & \cellcolor[HTML]{F87C7E}-47.34\% (±1.69\%) & \cellcolor[HTML]{F99B9E}-39.27\% (±2.56\%) \\ \hline
            $H_{max}$ (\texttt{ic.h\_max})      & \cellcolor[HTML]{BBCEE8}-8.56\% (±0.87\%)  & \cellcolor[HTML]{9FBBDF}-6.14\% (±0.65\%)  & \cellcolor[HTML]{FBF9FC}-15.08\% (±1.76\%) \\ \hline
            $M_0$ (\texttt{ic.m0})       & \cellcolor[HTML]{E2EAF6}-12.02\% (±1.02\%) & \cellcolor[HTML]{FBE2E5}-20.98\% (±1.60\%) & \cellcolor[HTML]{5A8AC6}-0.09\% (±0.52\%)  \\ \hline
        \end{tabular}
    \label{tab:hc_sorter:feature_permutations}
\end{table}

\section{Conclusion}\label{sec:conclusion}
This paper proposed the use of Hilbert space-filling curves in the context of optimisation problem landscape analysis for the purpose of (1) sampling the search space in a spatially correlated manner that also guarantees uniform coverage, and (2) spatially ordering samples generated using other  sampling algorithms such as the Latin hypercube. Experiments were conducted to evaluate the relative computational efficiency of the Hilbert curves, as well as the saliency of the landscape features extracted using Hilbert curve sampling and Hilbert curve ordering. In the context of sampling, Hilbert curves were significantly faster than Latin hypercube sampling for the purpose of generating order-sensitive features such as the information content metrics. Features extracted by the Hilbert curves were informative, and allowed for successful discrimination between problem classes. As an ordering tool, Hilbert curves performed significantly faster than the commonly used nearest neighbour ordering, and also yielded salient landscape features. Thus, Hilbert curves present a viable computationally efficient alternative to both Latin hypercube sampling, and nearest neighbour ordering of a sample for the purpose of landscape analysis.

%
%
%
\bibliographystyle{splncs04}
\bibliography{References}

\end{document}